\documentclass[runningheads]{llncs}
\usepackage[T1]{fontenc}

\usepackage{url}
\usepackage{hyperref}
\usepackage{hyperxmp}
\usepackage{xcolor}
\usepackage{tikz}
\usepackage{courier}
\usepackage{tabularx}
\usepackage{makecell}
\usepackage{threeparttable}
\usepackage{booktabs}
\usepackage{graphicx}  
\usepackage{amsmath}  
\usepackage{newtxmath}
\usepackage[numbers]{natbib}
\usepackage{mathtools}
\usepackage[stable]{footmisc}
\usepackage{color}

\begin{document}
%
\title{ProvenanceTrace: Atomic Claim Verification Across Financial and Open-Domain AI Systems}
\titlerunning{ProvenanceTrace: Claim-level Verification in AI Systems}
%

\author{Krati Saxena\inst{1}\orcidID{0000-0001-7049-9685} \and
Bhaskarjit Sarmah\inst{1}\orcidID{0009-0004-3076-9539} \and
Stefano Pasquali\inst{2}\orcidID{0009-0007-8005-3207}}
\authorrunning{K. Saxena et al.}
%
\institute{Domyn, Gurugram, India \and
Domyn, New York, USA
\email{krati.saxena,bhaskarjit.sarmah,stefano.pasquali@domyn.com}\\
\url{https://www.domyn.com/}}
\maketitle              
\begin{abstract}
Enterprise use of Large Language Models (LLMs) is expanding rapidly, but high-stakes deployment in enterprise information systems (EIS), especially in domains like finance, remains constrained by auditability and accountability requirements. Existing provenance systems are mostly document-level, leaving claim-level evidence gaps that increase review effort and governance risk. In this vision paper, we present \textbf{ProvenanceTrace} as an enterprise integration framework that decomposes outputs into atomic claims and links each claim to evidence (or flags it unsupported) through retrieval and verification. We present preliminary proof-of-concept (PoC) evidence intended to validate architectural feasibility and integration viability in both a financial agentic system and open-domain question answering. These early results suggest that claim-level provenance can operate across heterogeneous workflows, improve audit readiness through evidence links per claim, and support architectures integrated with governance, risk, and compliance processes.
\end{abstract}
\vspace{-3mm}
\section{Introduction}
\label{sect:intro}

Large Language Models (LLMs) are increasingly embedded in enterprise information systems (EIS) that support regulated reporting, compliance, and decision workflows in finance. From an Information Systems Engineering (ISE) perspective, this shifts the problem from model performance to governance of socio-technical decision processes: enterprises must demonstrate accountability, traceability, and control across models, data assets, and operational systems. Yet today’s RAG citations and system logs provide document-level traceability, not verification of the specific statements that drive action. This gap is material in finance, where regulations such as the EU AI Act, GDPR Article 22, and MiFID IIrequire auditable and accountable automation. 

Related work has advanced several pieces of the problem, including orchestration traces, lineage modeling, and fact-checking attribution \citep{LangChain2025, CrewAI2024, souza2025prov, han2024prov, abedin2022designing, liu2025olmotrace}. However, these strands are often disconnected in enterprise practice: lineage is captured without claim-level grounding, or claim checks are performed without workflow-level accountability integration. To address this gap, we position \textbf{ProvenanceTrace} as an enterprise integration blueprint that decomposes outputs into \emph{atomic claims} and links each claim to supporting evidence (or marks it unsupported), so provenance can operate as an ISE capability embedded in governance workflows through provenance-ready architectures, service-level controls, and governance, risk and compliance integration.

We provide an initial PoC to demonstrate feasibility and to anchor a research agenda for operationalization in real organizations. The contributions are as follows:
\begin{itemize}
    \item \textbf{Enterprise integration of claim-level provenance:} We show how atomic claim traceability supports accountability and governance in enterprise use.
    \item \textbf{Financial-domain application blueprint:} We present an end to end framework that decomposes outputs into claims and links them to evidence.
    \item \textbf{Early feasibility signals:} We report proof of concept results in a financial agentic workflow and open domain QA to motivate an ISE evaluation agenda.
\end{itemize}

\vspace{-5mm}
\section{Related Works}
\label{sect:rel_works}

Agentic AI systems require richer provenance than traditional pipelines. Recent provenance frameworks capture workflow execution well, including prompts, tool calls, and reasoning traces \citep{souza2025prov, yao2022react}, and reliability-oriented governance metrics for finance \citep{raza2025trism}. Earlier lineage systems such as PROV-IO+ and yProv \citep{han2024prov, abedin2022designing} improve reproducibility and interoperability, but they are not designed for atomic claim verification. As a result, most provenance approaches remain strong on process tracking but weak on claim-level auditability, which is central in regulated enterprise settings.

In parallel, claim verification and attribution methods have advanced retrieval-grounded checking and fine-grained citation alignment. Representative work includes RAGE for retrieval-linked verification \citep{rorseth2024towards}, OLMoTrace for span-level attribution \citep{liu2025olmotrace}, and SelfCite for sentence-level citation calibration \citep{Chuang2025SelfCiteSA}; long-form and agentic factuality pipelines further improve robustness \citep{wei2024long}. However, these methods are often evaluated as stand-alone fact-checking components rather than integrated with enterprise provenance infrastructure. ProvenanceTrace focuses on that integration point, not on claim verification itself: it unifies claim-centric verification with workflow-aware provenance so evidence, lineage, and accountability remain connected across financial agentic systems and open-domain question answering.

\begin{figure}[!t]
\centering
\includegraphics[width=0.75\textwidth]{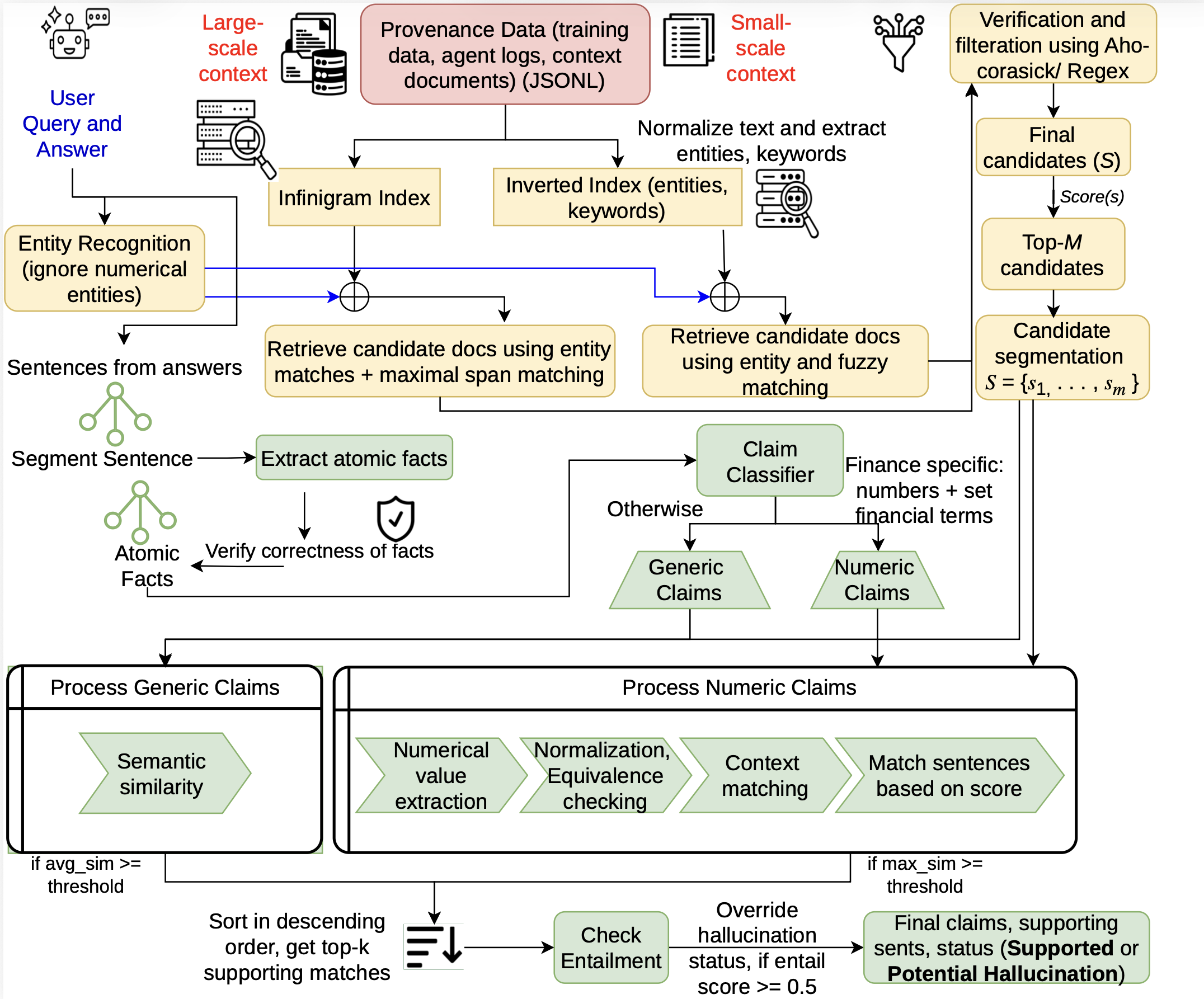}
\caption{Overview of ProvenanceTrace. Red, yellow and green components corresponds to \emph{Data Layer}, \emph{Index and Retrieval Layer} and \emph{Claim Verification Layer}, respectively.} 
\label{fig:overview}
\end{figure} 

\vspace{-3mm}
\section{Why Claim-Level Provenance Matters}
\label{sect:claimlevel_provenance}
Atomic-claim traceability turns AI outputs into accountable decision inputs by treating each claim as a verifiable unit with a status, linked evidence, and ownership. This enables organizations to justify approvals and audit why a decision was accepted, escalated, or blocked. The need is especially acute as faster, more autonomous AI workflows increase the risk that unsupported statements propagate into reports or customer-facing actions. We evaluate this idea in both financial agentic workflows and open-domain question answering, showing that claim-level accountability is useful across high-stakes and general-purpose settings.

\vspace{-3mm}
\section{Methodology}
\label{sect:methodology}
We propose ProvenanceTrace as a three-layer, enterprise-ready method for claim-level provenance, combining practical NLP components with governance-oriented design; \autoref{fig:overview} summarizes the framework.

\vspace{-3mm}
\subsection{Data Layer}
\label{subsect:data_layer}
The \emph{Data Layer} stores all source information needed to reconstruct provenance. In agentic AI workflows, this includes data fetched, generated, or referenced during execution, reasoning steps, tool outputs, and decision logs. In non-agentic LLM systems, it comprises the corpora and context documents used during inference. We use a standardized JSONL format so inputs from different sources are consistent: each line represents a provenance unit such as a document chunk, a tool response, or a workflow step with metadata. 

\vspace{-3mm}
\subsection{Index and Retrieval Layer}
\label{subsect:index_retrieval_layer}
This layer enables efficient retrieval through indexing optimized for dataset scale and computational constraints. For large corpora we use scalable substring indexing with Infini-gram \citep{Liu2024InfiniGram, liu2025olmotrace}; for smaller repositories we use a conventional inverted index over normalized text. We normalize text (lowercasing and whitespace cleanup) and extract named entities using SpaCy’s NER\footnote{\url{https://spacy.io/usage/linguistic-features\#named-entities}}; if entities are sparse, we add keywords via KeyBERT\footnote{\url{https://github.com/MaartenGr/KeyBERT}}. These entities/keywords act as anchors in the index, making retrieval fast and transparent to reviewers.

For each claim, we extract anchors from both the user query and the model answer, then retrieve candidate evidence from the chosen index. With Infini-gram, we use maximal matching spans to find strong substrings across large corpora; with the inverted index, we use exact matches and light fuzzy matching for abbreviations or typos. We also verify that anchors actually appear in the retrieved text (e.g., via Aho-Corasick or regex checks) to reduce false positives. We score the candidates using a simple blend of Jaccard overlap, syntactic similarity, and semantic similarity (e.g., SBERT\footnote{\url{https://sbert.net/}}), then rank them to produce a small, reviewable evidence set for each claim. This ranked evidence set is the supporting evidence used to verify the atomic claims in the next layer.

\vspace{-3mm}
\subsection{Claim Verification Layer}
\label{subsect:claim_verification}
The \emph{Claim Verification Layer} turns model output into an \emph{auditable set of atomic claims} with a corresponding evidence trail, so each claim can be supported, flagged for review, or blocked based on enterprise policy. It performs three steps that keep the method understandable while still audit-ready.

\textbf{Claim extraction.} Model responses often bundle multiple assertions in one sentence. For example, "In Q4 2025, the company reported revenue of USD 4.2B and net income of USD 0.8B" contains two atomic claims: "The company reported revenue of USD 4.2B in Q4 2025" and "The company reported net income of USD 0.8B in Q4 2025". We split responses into sentences and extract minimal, self-contained claims that can be verified independently. We use a lightweight two-stage generation and verification process: first generate candidate claims with a fixed enterprise LLM to reduce operational variance, then validate each claim against the source sentence to ensure clarity and atomicity. We remove the duplicates and keep track of claim positions for traceability. Parallely, we normalize, de-duplicate and segment the candidates obtained from the retrieval layer above. 

\textbf{Policy-driven routing and alignment.} To support reliable decisions and auditability in EIS workflows, ProvenanceTrace routes claims through different verification paths based on policy: generic text claims follow a semantic support check against source evidence, while numeric or finance-critical claims go through stricter validation. For numeric claims, the system standardizes number formats and structured representations (e.g., key--value records), then checks value, unit, and scale consistency before assigning support. The final claim--evidence alignment combines structural match, domain terminology, and contextual meaning to produce a reviewable decision trail. Finally, any claims initially flagged as unsupported or potential hallucinations undergo a definitive entailment analysis using a specialized language model, ensuring that all synthesized outputs maintain the strict factual integrity required for enterprise deployment.

\vspace{-3mm}
\section{Evaluation: Initial Feasibility Evidence}
\label{sect:experiments_results}
\vspace{-3mm}
In this paper, we focus on reporting implemented proof-of-concept results (retrieval quality, hallucination routing behavior, and runtime/stability) across two representative settings to assess feasibility rather than benchmark leadership. 

\noindent\textbf{Financial Agentic Veracity (FAV):} This use-case evaluates provenance tracing in a realistic enterprise financial workflow where a central orchestrator coordinates 8 specialized agents (web searchers, data scrapers, summarizers, extractors, analysts, and advisors) to turn market information into investment recommendations; in practice, this reflects how organizations combine multiple services before a decision is issued. For a query such as "Should I buy this stock based on its P/E ratio and recent earnings?", agents gather market/public evidence, synthesize findings, extract structured facts, and produce advice, while agent logs act as data-layer artifacts for auditability. Provenance on agentic data is especially important because final outputs are composed from multiple intermediate tool calls and transformations, so accountability requires tracing both the final claim and the execution path that produced it. The dataset combines templated financial queries with curated ticker symbols across Technology and Financial sectors, covering investment decisions, market analysis, and portfolio management; we run 420 unique prompts twice (840 total runs).

\noindent\textbf{Open-Domain Wikipedia Fact Audit (OWFA):} This use-case uses the "allenai/olmo-mix-1124" Wikipedia dataset\footnote{\url{https://huggingface.co/datasets/allenai/olmo-mix-1124/viewer/wiki}} as a trusted reference and includes questions such as "In what year did World War II begin?" and "Which organization is responsible for the Nobel Prize?" We run 241 prompts three times each (723 total runs). Beyond the finance-specific case, this open-domain evaluation provides preliminary indications that the same architecture can be applied to larger, heterogeneous corpora common in enterprise environments. 

\noindent\textbf{Evaluation Metrics:}
We evaluate ProvenanceTrace on three enterprise-relevant feasibility dimensions: retrieval accuracy, hallucination detection, and operational robustness. Retrieval is measured by Hit@5 (at least one supporting source in the top five), average first-relevant rank, and full rank distributions, using an LLM judge for supportiveness; hallucination detection checks claim support via semantic similarity and LLM-based entailment, reporting score distributions, means, and hallucination counts. For scalable PoC evaluation at high claim volume, we use Domyn-Small\footnote{\url{https://www.domyn.com/language-models}} (temperature = 0.0) with fixed prompts for Hit@5 and support labeling (released for reproducibility\footnote{\url{https://github.com/kratisaxena-domyn/ProvenanceTrace_Assets}}), and assess robustness through end-to-end runtime plus stability (coefficient of variation) over answer length, fact counts, and support labels across runs. This submission reports preliminary feasibility evidence and does not yet include comparative baselines, component ablations, human calibration studies, or sensitivity analysis over key thresholds.

\vspace{-3mm}
\subsection{Results}
\label{subsect:results}

\begin{table}[t!]
\centering
\caption{Consolidated Results for FAV and OWFA Experiments}
\label{tab:consolidated_results}
{\small
\setlength{\tabcolsep}{3pt}
\begin{tabular*}{\linewidth}{@{\extracolsep{\fill}}llcc@{}}
\toprule
\textbf{Category} & \textbf{Metric} & \textbf{FAV} & \textbf{OWFA} \\
\midrule
\textit{Retrieval} & Hit@5 Success Rate (\%) & 91.3 & 72.7 \\
\textit{Retrieval} & Mean Rank (Successful Hits) & 1.4 & 1.7 \\
\textit{Hallucination} & Total Claims Evaluated & 23,093 & 3,082 \\
\textit{Hallucination} & Supported Rate (\%) & 87.4 & 69.6 \\
\textit{Hallucination} & Hallucination Rate (\%) & 12.6 & 30.4 \\
\textit{Efficiency/Stability} & Mean Runtime (s) & 84.71 & 161.03 \\
\textit{Efficiency/Stability} & Answer Length CV & $\sim$0.4 (moderate) & $\sim$0.2 (lower) \\
\textit{Efficiency/Stability} & Fact Count CV & $\sim$0.4 (moderate) & $\sim$0.8--1.2 (broad) \\
\textit{Efficiency/Stability} & Support Rate CV & near zero & $\sim$1.0 (broad) \\
\bottomrule
\end{tabular*}
}
\end{table}

\textbf{Hit@5 Analysis}: As shown in Table~\ref{tab:consolidated_results}, the stronger FAV retrieval results are expected in this PoC by design, not a weakness in high-stakes assurance. The FAV pipeline includes domain-specific financial normalization and numeric consistency checks, which improve matching for structured enterprise evidence. OWFA, in contrast, uses a more generic semantic verification setup over heterogeneous open-domain content, where lexical and contextual variation naturally introduces more misses. For information systems, this demonstrates that claim-level provenance can be tuned to domain risk and data characteristics: adding domain-grounded controls (especially numeric checks) improves audit performance while preserving the same core provenance architecture.

\textbf{Hallucination Detection Analysis}: As shown in Table~\ref{tab:consolidated_results}, FAV has lower hallucination rates and higher support rates because its evidence space is smaller and tighter with agent data and logs, making the top-$M$ set (here $M=5$) more likely to contain directly supporting evidence. This aligns with the higher Hit@5 in FAV. OWFA uses a much larger and noisier Wikipedia space, where top-$M$ candidates are often only partially relevant, reducing Hit@5 and support rates. For information systems, this is expected in the current PoC design and indicates clear improvements: adaptively increase $M$ and improve candidate selection to better capture relevant evidence spans. Statistical separation remains strong in both settings: similarity and entailment distinguish supported from unsupported claims with p-values below 0.0001 in t-tests and Mann--Whitney tests.

\textbf{General Analysis and Runtime Insights}:
System efficiency and stability show clear domain differences. In FAV, mean runtime is 84.71 seconds and is tightly clustered, with most cost coming from multi-agent orchestration. In OWFA, mean runtime rises to 161.03 seconds, driven mainly by hallucination scoring and consistency alignment over large, unstructured Wikipedia evidence. Coefficient of variation analysis reinforces this contrast. In FAV, answer length and fact count vary moderately, while support rate CV stays near zero, indicating stable verification consistency. In OWFA, constrained prompts reduce answer-length variation, but heterogeneous evidence causes broader spread in fact count and support rate. The broader OWFA CV spread is statistically significant (p-value < 0.01) and reflects greater extraction and claim-support variability.

ProvenanceTrace, as a focused proof-of-concept for enterprise and financial-domain use of atomic claim-level provenance, prioritizes architecture feasibility over exhaustive benchmark comparison. We leave direct baseline and attribution comparisons (e.g., BM25+cross-encoder \citep{lu2025pathwayrelevancecrossencodersimplement}, OLMoTrace \citep{liu2025olmotrace}, and SelfCite \citep{Chuang2025SelfCiteSA}) to future integration studies.

\vspace{-3mm}
\section{Discussion}
\label{sect:discussion}

In this vision paper, we present ProvenanceTrace as a practical step toward provenance-ready enterprise systems. Instead of treating an LLM response as a single opaque output, it treats each atomic claim as an auditable unit linked to concrete evidence. This shift is important because organizations do not govern answers in aggregate; they govern decisions, controls, and accountability at claim level.

This is especially relevant in the current AI landscape, where LLMs are moving from assistive drafting to workflow participation in reporting, compliance, and customer-facing operations. In these settings, document-level citations are often insufficient for sign-off, escalation, and post-hoc audit. Claim-level provenance fills this gap by connecting generated statements to evidence in a form that can be used by risk, compliance, and audit functions.

For organizations, adoption can be incremental rather than disruptive. A practical rollout path is to start in one high-value workflow, define provenance SLAs (coverage, latency, escalation), and embed verification checkpoints into existing processes. The PoC results suggest that the architecture can be adapted to structured enterprise contexts and broader domains with appropriate retrieval controls and review policies.

Open challenges remain, but they are now concrete engineering and governance tasks rather than abstract trust concerns. Next steps are to complement LLM-based judging with calibrated human review under explicit policies, add retrieval and attribution baselines, and report clearer quality measures for claim extraction and atomicity. We will also specify dataset and prompting protocols more explicitly, deepen numeric-verification details for unit/scale/currency edge cases, and formalize governance integration through standards-aligned provenance mappings and escalation schemas. These steps strengthen external validity while preserving the paper’s vision focus on provenance-ready enterprise architectures.

\begin{credits}

\subsubsection{\discintname}
The authors have no competing interests.
\end{credits}
%
%
%
%

\bibliographystyle{splncs04}
\bibliography{references}

@inproceedings{souza2025prov,
  title={PROV-AGENT: Unified provenance for tracking AI agent interactions in agentic workflows},
  author={Souza, Renan and Gueroudji, Amal and DeWitt, Stephen and Rosendo, Daniel and Ghosal, Tirthankar and Ross, Robert and Balaprakash, Prasanna and Da Silva, Rafael Ferreira},
  booktitle={2025 IEEE International Conference on eScience (eScience)},
  pages={467--473},
  year={2025},
  organization={IEEE}
}

@article{raza2025trism,
  title={Trism for agentic ai: A review of trust, risk, and security management in llm-based agentic multi-agent systems},
  author={Raza, Shaina and Sapkota, Ranjan and Karkee, Manoj and Emmanouilidis, Christos},
  journal={arXiv preprint arXiv:2506.04133},
  year={2025}
}

@article{abedin2022designing,
  title={Designing and managing human-AI interactions},
  author={Abedin, Babak and Meske, Christian and Junglas, Iris and Rabhi, Fethi and Motahari-Nezhad, Hamid R},
  journal={Information Systems Frontiers},
  volume={24},
  number={3},
  pages={691--697},
  year={2022},
  publisher={Springer}
}

@inproceedings{rorseth2024towards,
  title={Towards Explainability in Retrieval-Augmented LLMs},
  author={Rorseth, Joel and Godfrey, Parke and Golab, Lukasz and Srivastava, Divesh and Szlichta, Jaroslaw},
  booktitle={2024 IEEE 40th International Conference on Data Engineering (ICDE)},
  pages={5669--5670},
  year={2024},
  organization={IEEE}
}

@inproceedings{yao2022react,
  title={React: Synergizing reasoning and acting in language models},
  author={Yao, Shunyu and Zhao, Jeffrey and Yu, Dian and Du, Nan and Shafran, Izhak and Narasimhan, Karthik R and Cao, Yuan},
  booktitle={The eleventh international conference on learning representations},
  year={2022}
}

@article{han2024prov,
  title={PROV-IO++: A Cross-Platform Provenance Framework for Scientific Data on HPC Systems},
  author={Han, Runzhou and Zheng, Mai and Byna, Suren and Tang, Houjun and Dong, Bin and Dai, Dong and Chen, Yong and Kim, Dongkyun and Hassoun, Joseph and Thorsley, David},
  journal={IEEE Transactions on Parallel and Distributed Systems},
  volume={35},
  number={5},
  pages={844--861},
  year={2024},
  publisher={IEEE}
}

@article{liu2025olmotrace,
  title={OLMoTrace: Tracing Language Model Outputs Back to Trillions of Training Tokens},
  author={Liu, Jiacheng and Blanton, Taylor and Elazar, Yanai and Min, Sewon and Chen, YenSung and Chheda-Kothary, Arnavi and Tran, Huy and Bischoff, Byron and Marsh, Eric and Schmitz, Michael and others},
  journal={arXiv preprint arXiv:2504.07096},
  year={2025}
}

@article{Chuang2025SelfCiteSA,
  title={SelfCite: Self-Supervised Alignment for Context Attribution in Large Language Models},
  author={Yung-Sung Chuang and Benjamin Cohen-Wang and Shannon Zejiang Shen and Zhaofeng Wu and Hu Xu and Xi Victoria Lin and James Glass and Shang-Wen Li and Wen-tau Yih},
  journal={ArXiv},
  year={2025},
  volume={abs/2502.09604},
  url={https://api.semanticscholar.org/CorpusID:276317486}
}

@article{wei2024long,
  title={Long-form factuality in large language models},
  author={Wei, Jerry and Yang, Chengrun and Song, Xinying and Lu, Yifeng and Hu, Nathan and Huang, Jie and Tran, Dustin and Peng, Daiyi and Liu, Ruibo and Huang, Da and others},
  journal={Advances in Neural Information Processing Systems},
  volume={37},
  pages={80756--80827},
  year={2024}
}

@article{Liu2024InfiniGram,
  title={Infini-gram: Scaling Unbounded n-gram Language Models to a Trillion Tokens},
  author={Liu, Jiacheng and Min, Sewon and Zettlemoyer, Luke and Choi, Yejin and Hajishirzi, Hannaneh},
  journal={arXiv preprint arXiv:2401.17377},
  year={2024}
}

@misc{LangChain2025,
  author = {{LangChain Inc.}},
  title = {LangChain: A platform for reliable agents},
  year = {2025},
  note = {Available at https://www.langchain.com/ (Accessed: 2025-11-14)}
}

@misc{CrewAI2024,
  title        = {CrewAI: A multi-agent automation framework},
  author       = {{CrewAI Inc.}},
  year         = {2024},
  note         = {Available at https://www.crewai.com/ (Accessed: 2025-11-14)}
}

@misc{lu2025pathwayrelevancecrossencodersimplement,
      title={Pathway to Relevance: How Cross-Encoders Implement a Semantic Variant of BM25}, 
      author={Meng Lu and Catherine Chen and Carsten Eickhoff},
      year={2025},
      eprint={2502.04645},
      archivePrefix={arXiv},
      primaryClass={cs.IR},
      url={https://arxiv.org/abs/2502.04645}, 
}
%




\end{document}